\documentclass[letterpaper, 10 pt, conference]{ieeeconf}  % Comment this line out if you need a4paper

\IEEEoverridecommandlockouts                              % This command is only needed if 
\usepackage{setspace} % if you already have it
\usepackage{amsmath}
\usepackage{makecell} % in preamble
\usepackage[backend=biber,style=ieee]{biblatex}
\usepackage{bm}

\usepackage{multirow}
\usepackage{graphicx} % Required for including images
\usepackage{booktabs} % Recommended for professional-looking tables
\usepackage{xcolor}

\definecolor{tan}{rgb}{0.82, 0.71, 0.55}
  \usepackage{hyperref}
\title{\LARGE \bf
EmbodiSwap for Zero-Shot Robot Imitation Learning
}

\usepackage{capt-of}   % lightweight \captionof

 \author{Eadom Dessalene*, Pavan Mantripragada*, Michael Maynord and Yiannis Aloimonos% <-this % stops a space
\thanks{* These authors contributed equally to this work.}% <-this % stops a space
\thanks{The authors are with the department of Computer Science, University of
Maryland, College Park, MD, 20742}
}

\begin{document}

\maketitle
\thispagestyle{empty}
\pagestyle{empty}

% % \clearpage
% % \begin{figure*}[H]
% % \begin{figure*}[!htb]
% \begin{center}
%     \includegraphics[width=1\textwidth]{Figure_1_Architecture_21.png}

% \centering
%     \caption{\textbf{Overview of our training setup}: Our system takes a sequence of frames $\{I_0, ..., I_T\}$ featuring a human actor performing an action as input. The first frame of this sequences is passed to a multi-step \textbf{Robot Compositing} process, producing an image $I^*_0$ with the human hand of $I_0$ substituted with a robot manipulator. The robot image $I^*_0$ is passed into \textbf{V-JEPA encoder}. The output of the encoder is passed along with a stack of positional mask tokens $M_{1:T}$ that correspond to frames $I_{1:T}$ to the \textbf{V-JEPA predictor}. The output of the predictor is then fed along with optional (encoded by dashed lines) encoded representations of proprioception token $p_0$ and an action location token $l_0$ (both associated with $I_0$) into cross-attention layers $C$. $C$ produces as output a 9 DOF relative hand transform prediction, corresponding to a relative predicted hand transform from $I_0$ to $I_T$. Training is supervised using $L_1$ loss with a training signal of a relative 3D transform of the hand as derived between images $I_0$ and $I_T$ by the \textbf{3D Hand Reconstruction} network. For clarity, model and 3D Hand Reconstruction output are overlayed onto $I_T$ as output in the figure.}
    
%     \label{figure_architecture}
% % \end{figure*}
% \end{center}
% % \clearpage

%%%%%%%%%%%%%%%%%%%%%%%%%%%%%%%%%%%%%%%%%%%%%%%%%%%%%%%%%%%%%%%%%%%%%%%%%%%%%%%%
\begin{abstract}

%phantom - collected manually (differs in data)
%zero mimic - uses goal images (differs in model)
%masquerade - not zero-shot (differs in task)
% produces an average $35\%$ and $21\%$ reduction in test error of end-effector translation and rotation predictions, respectively, in comparison to 
We introduce \textbf{EmbodiSwap} - a method for producing photorealistic synthetic robot overlays over human video. We employ EmbodiSwap for zero-shot imitation learning, bridging the embodiment gap between in-the-wild ego-centric human video and a target robot embodiment. We train a closed-loop robot manipulation policy over the data produced by EmbodiSwap. We make novel use of V-JEPA as a visual backbone, repurposing V-JEPA from the domain of video understanding to imitation learning over synthetic robot videos. Adoption of V-JEPA outperforms alternative vision backbones more conventionally used within robotics. In real-world tests, our zero-shot trained V-JEPA model achieves an $82\%$ success rate, outperforming a few-shot trained $\pi_0$ \cite{black2024pi0} network as well as $\pi_0$ trained over data produced by EmbodiSwap. We release (i) code for generating the synthetic robot overlays which takes as input human videos and an arbitrary robot URDF and generates a robot dataset, (ii) the robot dataset we synthesize over EPIC-Kitchens, HOI4D and Ego4D, and (iii) model checkpoints and inference code, to facilitate reproducible research and broader adoption.

\end{abstract}

% \begin{center}
%   \includegraphics[width=0.8\textwidth]{your-figure}
%   \captionof{figure}{Your caption here}
% \end{center}

%%%%%%%%%%%%%%%%%%%%%%%%%%%%%%%%%%%%%%%%%%%%%%%%%%%%%%%%%%%%%%%%%%%%%%%%%%%%%%%%
\section{INTRODUCTION}

% % {\color{blue} (All "action prediction" etc. go to "imitation learning")}

% {\color{blue} (TODO: tweak superiority of video prediction as a pre-training method to superiority of V-JEPA)}

% Ideas for paper titles and method acronym:

% EmbodiSwap-V-JEPA: Human-to-Robot from Internet Video, without Robot Demos

% Zero-Shot V-JEPA with EmbodiSwap: blahblah (without Robot Demos)

% Zero-Shot V-JEPA with EmbodiSwap: Human-to-Robot Action from Internet Video, without Robot Demos (maybe remove Internet video)

% ({\color{red} To Pavan}: subsections below are just for structuring writing, they will be removed when flow is added between subsections)

% \subsection{1: Zero-Shot Framing}

% {\color{blue} (TODO: make it clear that the task is action prediction, not classification)}

Acquiring robot demonstrations for every task, environment, and embodiment is prohibitively expensive, while human video is abundant and easier to gather. Leveraging this, our zero-shot imitation learning approach enables robots to perform new actions without robot demonstrations or goal images, deriving training supervision solely from in-the-wild human video, unlike \cite{lepert2025phantom} which depends on curated in-lab demonstrations.

Traditional robot learning generally relies on demonstrations collected via teleoperation, VR \cite{higuchi2024end}, or scripted control. Real-world datasets are costly, hardware- and environment-specific, and often severely biased (see Figure 2 in \cite{hoque2025egodex}). In contrast, human videos are abundant, diverse, and naturally rich in hand–object interactions. Leveraging large-scale human video not only circumvents the challenges of robot data collection but also broadens the range of actions robots can learn to perform \cite{jaquier2025transfer}.

% \begin{figure*}[H]
\begin{figure*}[!htb]
    \includegraphics[width=1\textwidth]{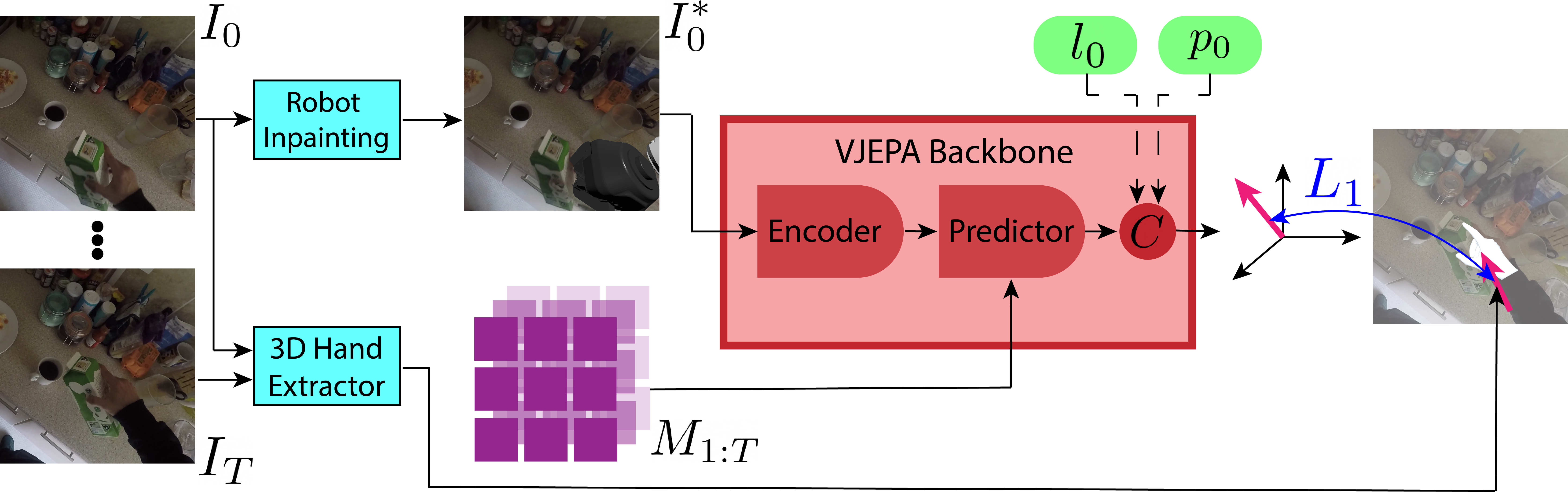}

\centering
    \caption{\textbf{Overview of our training setup}: Our system takes a sequence of frames $\{I_0, ..., I_T\}$ featuring a human actor performing an action as input. The first frame of this sequences is passed to a multi-step \textbf{Robot Compositing} process, producing an image $I^*_0$ with the human hand of $I_0$ substituted with a robot manipulator. The robot image $I^*_0$ is passed into \textbf{V-JEPA encoder}. The output of the encoder is passed along with a stack of positional mask tokens $M_{1:T}$ that correspond to frames $I_{1:T}$ to the \textbf{V-JEPA predictor}. The output of the predictor is then fed along with optional (represented by dashed lines) encoded representations of proprioception token $p_0$ and an action location token $l_0$ (both associated with $I_0$) into cross-attention layers $C$. $C$ produces as output a relative hand transform prediction, corresponding to a relative predicted hand transform from $I_0$ to $I_T$. Training is supervised using $L_1$ loss with a training signal of a relative 3D transform of the hand as derived between images $I_0$ and $I_T$ by the \textbf{3D Hand Reconstruction} network. For clarity, model and 3D Hand Reconstruction output are overlayed onto $I_T$ as output in the figure.}
    
    \label{figure_architecture}
\end{figure*}

% % \clearpage

% \subsection{3: Characterization of Data Produced and Released}

As robot demonstrations are hard to scale, and large scale human data lacks robot embodiments, we introduce a method - EmbodiSwap - and associated data, which swaps the human embodiment in human video with a robot embodiment. We release a robot manipulation dataset consisting of a large number of robot overlays on egocentric video, \href{https://drive.google.com/drive/folders/1-UUywelBCOe-E_ErpoaAHQa4dgjq6AfH?usp=sharing}{here}, focusing on its utility for zero-shot imitation learning, though it can support other tasks such as few-shot or inverse RL.

% While such data could be used in a number of settings such as few-shot imitation learning or inverse reinforcement learning, we focus on its utility for zero-shot imitation learning and release it for broader use.

% While such data could be used generically for a range of action understanding tasks,

% While such data could be used generically for a range of action undertsanding approaches,

% While such data could be used to learn generalizable pre-trained representations for a few-shot imitation learning setting, we focus on its utility for zero-shot imitation learning and release it for broader use.

% While such data could serve a range of applications—including zero-shot imitation learning or pre-trained representation learning—we focus on its utility for zero-shot imitation learning and release it for broader use.

% This pipeline yields training data that combines the scale and diversity of human video with the embodiment of robot demonstrations, enabling direct use for imitation learning.

EmbodiSwap transforms human egocentric videos into \textit{robot composited} demonstrations through a multi-step video-editing process. Hand reconstruction networks recover camera-compensated 3D hand trajectories and produce high-resolution actor masks, which are passed to an inpainting model to remove the human actor and their effects. A photorealistic, pose-aligned robotic hand is composited into the scene using hallucinated depth maps, seamlessly replacing the human actor. Each robot frame is paired with the future end effector pose that drives the action, and these labels are included in the dataset.

 % the end-effector is to go \textit{next} in performing the action.

% at the pose the end effector will be at the set offset into the future

% \subsection{4: Architecture Setup}
% We apply this data to learn closed loop robot manipulation policies through zero-shot imitation learning. At the core of our system is a closed loop video-predictive transformer policy network built on V-JEPA \cite{bardes2023v}. We employ V-JEPA weights pretrained on a large-scale video dataset of over 2 million diverse clips of humans performing actions. V-JEPA is a video transformer architecture consisting of a separate pre-trained encoder network and predictor network. We freeze the encoder and fine-tune only the predictor and a lightweight attention probe. Given an input frame, the encoder produces a representation passed to the predictor. The output of the predictor is concatenated with optional proprioception and optional action location tokens, which is subsequently fed to an attention probe which predicts the robot hand's future pose, supervised via L1 loss against true hand alignment. 

We use this data to train closed-loop robot manipulation policies for zero-shot imitation learning. At the core is a video-predictive transformer policy network built on V-JEPA \cite{bardes2023v}, pretrained over 2M human action clips. We freeze V-JEPA's encoder and fine-tune its predictor and lightweight attention probe. The encoder outputs a representation for each input frame; the predictor combines this with optional proprioception and action tokens, and the probe predicts the robot hand’s future pose, supervised with L1 loss.

During real-world inference we run the policy network in a closed loop: at each step the network outputs an action, the robot executes it, the environment changes, and the new observation is fed back to the network, repeated for a fixed number of steps. Note that during inference our model does not rely on auxiliary goal image conditioning unlike \cite{shi2025zeromimic}.

% Prior works that depend on goal images \cite{shi2025zeromimic} often require human involvement to collect those images - this both increases the labor burden, and weakens the “zero-shot” claim as the goal image itself is a partial demonstration of the task. By contrast, our framework avoids these limitations entirely. 

A considerable body of work has focused on learning robot policies from robot demonstrations \cite{team2024octo, black2024pi0}. While these methods make good generalist policies capable of following language instructions, they still require in-domain robot data for fine-tuning for new robotic platforms and unseen environments - in contrast our method does not.

% In contrast, our method leverages only egocentric human videos and requires no expert demonstrations, and demonstrates zero-shot transfer to novel scenes.

\begin{figure}[b!]
    \includegraphics[width=0.485\textwidth]{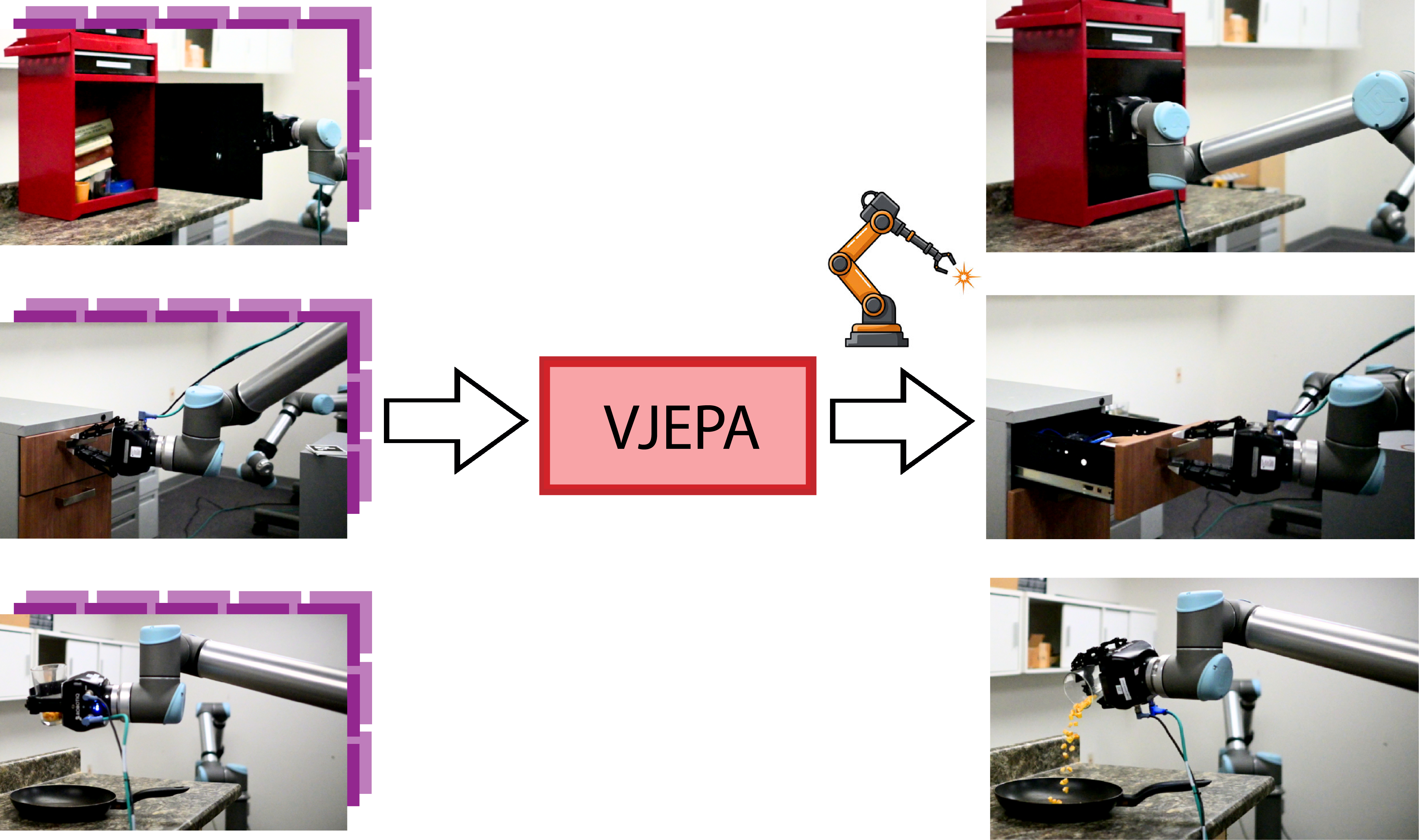}

    \caption{\textbf{Overview of our Zero-Shot test setup:} After training V-JEPA over distributions of scenes, objects, and embodiments, we deploy the network on \textit{out-of-distribution} embodiments, objects and environments. Test sequences — consisting of the first RGB frame plus positional masks for subsequent frames — are provided as input, producing as output a relative transformation predictions. We illustrate three example sequences, one each for: \textit{close}, \textit{open}, and \textit{pour}. Input is shown on the left, and the consequence of action execution is shown on the right.}
    \label{figure_deployment}
\end{figure}

\begin{figure*}[t!]
    \includegraphics[width=1\textwidth]{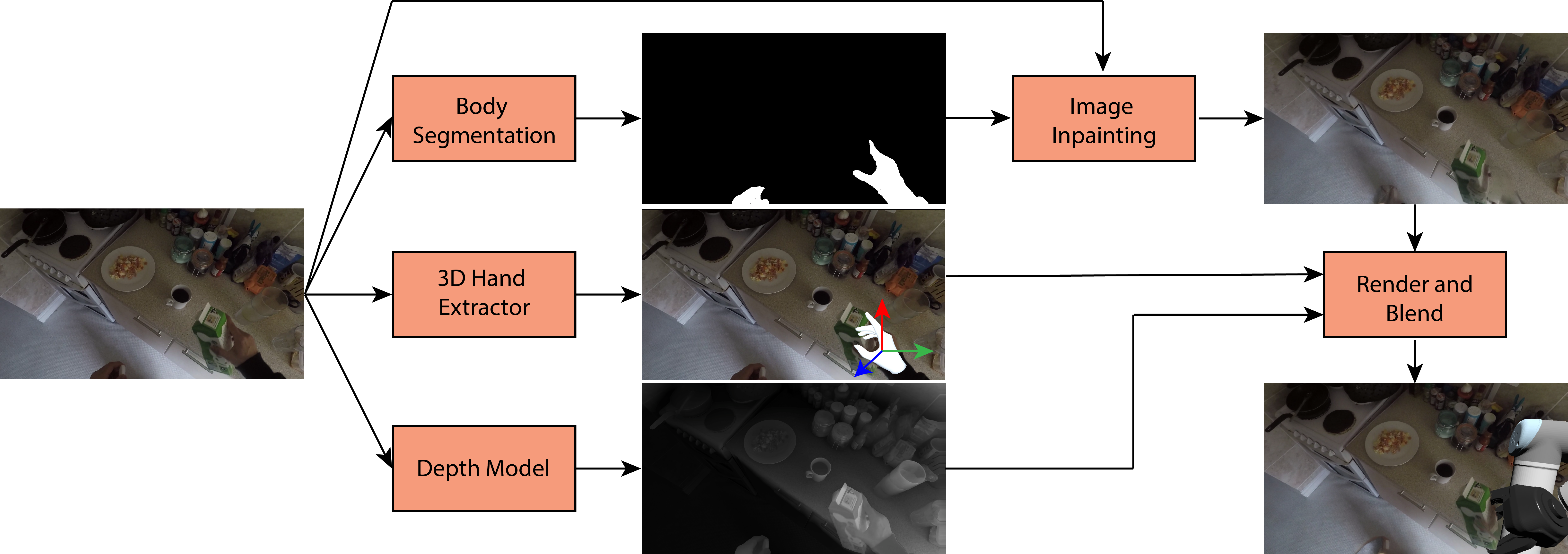}
    \centering
\caption{\textbf{Overview of our robot compositing pipeline:} The process begins with a human RGB frame in which a hand is visible. This image is processed first by three components: 1) a \textbf{Body Segmentation Network}, which produces a binary segmentation mask of the human actor; 2) a \textbf{3D Hand Extractor}, which reconstructs the human hand skeleton in 3D; and 3) a \textbf{Depth Network}, which estimates accurate metric depth (composite of grayscaled image and depth image shown for visualization purposes). The output of these components is then further processed by an additional two components: 4) \textbf{Image Inpainting}, which takes the original RGB image and the body segmentation and erases the human actor and their effects from the scene. And, finally 5) \textbf{Render and Blend}, which takes all of the inpainted image, depth map, and the end-effector pose, and renders a synthetic robot manipulator, composites it into the scene, and adjusts foreground/background contents based on the depth differences between the scene and the robot.}

    \label{figure_data_generation_flow}
\end{figure*}

We perform an extensive comparison across $13$ different pre-trained vision backbones. We observe that V-JEPA pre-trained for video feature-level prediction outperforms all other pre-training methods commonly used within robotics in addition to other non-robotics pre-training methods. 

% Furthermore, we perform real-world experimentation 

% Common pre-training methods in robotics include: 1) Internet-scale static image pretrained models trained via classification, 2) Internet-scale static image pretrained models trained for segmentation, and 3) recently released large-scale robot datasets used to learn useful visual representations. 

The primary contributions of this work are as follows:
\begin{itemize}
    % \item Novel use of Internet-scale video predictive models towards zero-shot imitation learning. We repurpose V-JEPA, producing a closed-loop manipulation policy, deploying in a zero-shot setting achieving an $82\%$ real-world success rate, outperforming across the board a few-shot trained $\pi_0$-FAST \cite{pertsch2025fast} model which has access to in-lab robot demonstrations, as well as the zero-shot $\pi_0$-FAST model trained over data produced by EmbodiSwap (which outperforms the few-shot trained $\pi_0$-FAST model).

    \item Novel use of Internet-scale video predictive models towards zero-shot imitation learning. We repurpose V-JEPA, producing a closed-loop manipulation policy, which achieves an $82\%$ real-world success rate over $5$ actions. Across all actions we outperform a few-shot trained $\pi_0$ model which has access to in-lab robot demonstrations, as well as the zero-shot $\pi_0$ model trained over data produced by EmbodiSwap (which outperforms the few-shot trained $\pi_0$ model).

    % \item Empirical demonstration of the superiority of using video prediction (a non-standard pre-training method in robot manipulation) over other pre-training methods, for the purpose of zero-shot robot manipulation learning. 

    \item Empirical demonstration of the superiority of using a V-JEPA model pre-trained for feature-level video prediction (a non-standard pre-training method in robot manipulation) over other pre-training methods, for the purpose of forecasting future end-effector trajectories. 
    
    % {\color{blue} (Quantify superiority of V-JEPA)}
    
    % \item Overlay of robot embodiment over existing egocentric video creating a large-scale robot manipulation dataset, as well as temporal boundaries well-suited for robot manipulation tasks as opposed to action classification tasks. We release all training data ({\color{blue}here}), checkpoints ({\color{blue}here}) and inference code ({\color{blue}here}). %release data annotations
    \item \textbf{EmbodiSwap:} Overlay of robot embodiment over existing egocentric video, as well as temporal boundaries well-suited for robot manipulation tasks, creating a large-scale robot manipulation dataset. We release all training data, checkpoints and inference code \href{https://drive.google.com/drive/folders/1-UUywelBCOe-E_ErpoaAHQa4dgjq6AfH?usp=sharing}{here}. Given a robot URDF and egocentric videos, we release the code for generating robot-composited videos.
    
    % \item Demonstration of zero-shot generalization to the real world, deploying a UR10 to perform $5$ different actions learned from data produced by EmbodiSwap, demonstrating an X\% improvement over OpenPi with limited access to real world robot demonstrations.

    % \item We deploy our model on a UR10 to perform $5$ different actions in a zero-shot setting achieving an $82\%$ real-world success rate, outperforming across the board a few-shot trained $\pi_0$ \cite{black2410pi0} model with access to in-lab robot demonstrations, as well as the zero-shot $\pi_0$ model trained over data produced by EmbodiSwap (which outperforms the few-shot trained $\pi_0$ model).

    % \item

%(as opposed to action classification tasks)

\end{itemize}

The rest of this paper is structured as follows: In Section \ref{sec_related_works} we detail related work, in Section \ref{sec_methods} we detail our method, in Section \ref{sec_experiments} we cover experiments, and in Section \ref{sec_conclusion} we conclude.

\section{RELATED WORK}
\label{sec_related_works}

\subsection{Forecasting Future Hand Trajectories}
% The problem of forecasting hand trajectories into the future is extremely challenging yet crucial for the alignment of actions between humans and robots. Among the first of this underexplored area of research, the work of \cite{dessalene2021forecasting} introduces \textit{contact anticipation maps}, or per-pixel estimates of time-to-contact between hands and objects belonging to the scene. The work of \cite{liu2022joint} predicts future hand motion in the form of 2D image coordinates by relying on synergies between hand trajectories and object affordances. The state of the art work of \cite{ma2025novel}
% develops this direction further, introducing explicit head motion forecasting as well as additional multimodal modalities. To our knowledge, there is no comparison of state-of-the-art vision backbones and their respective performances on the task of regressing future 6D hand poses.  This is important to explore as it determines the extent to which advancements in vision can be transferred directly into robot learning. We contribute an evaluation over $13$ different models, demonstrating the superiority of the V-JEPA backbone \cite{bardes2023v}.

Forecasting future hand trajectories is a difficult but essential problem for aligning human and robot actions. Early work by \cite{dessalene2021forecasting} introduced \textit{contact anticipation maps}, per-pixel time-to-contact estimates between hands and objects. \cite{liu2022joint} predicted 2D hand motion by modeling synergies between hand trajectories and object affordances. More recently, \cite{ma2025novel} utilizes explicit head motion prediction and additional modalities. However, no prior work has systematically compared state-of-the-art vision backbones for regressing future hand poses. We address this gap by evaluating $13$ models, demonstrating the superiority of V-JEPA \cite{bardes2023v}.

\subsection{Learning Robot Policies from Human Videos}
% The field of leveraging the scale of human video datasets for the learning of robot policies has been exploding. The majority of works \cite{singh2025hand,lin2024data,yang2024spatiotemporal} leverage human datasets for pre-training visual representations that require fine-tuning over a reduced (yet often still sizable) number of robot demonstrations due to the gains in sample efficiency and generalization. A few works as of recent \cite{shi2025zeromimic,liu2025egozero, lepert2025phantom}
% learn \textit{zero-shot} policies over human videos, requiring no additional robot demonstrations. Our work differs from \cite{shi2025zeromimic} in that while their approach requires a goal image, which itself is a partial demonstration of the action to be performed, our network does not rely on a goal image, and additionally addresses the human-to-robot embodiment gap that \cite{shi2025zeromimic} sidesteps. Our work additionally differs from \cite{liu2025egozero,lepert2025phantom} in that the data we train over is in-the-wild and therefore a far more challenging learning problem due to the enormous diversity in the styles and dynamics of the actions performed, as opposed to \cite{liu2025egozero,lepert2025phantom} in which the datasets are collected by the authors of the respective works. Very recent work \cite{lepert2025masquerade} concurrently submitted to ICRA 2026 is similar to our work, but is not zero-shot as their method requires additional real-world robot demonstrations to be acquired.

The use of large-scale human video datasets for robot policy learning has grown rapidly. Most works \cite{singh2025hand,lin2024data,yang2024spatiotemporal} leverage such data for visual pre-training, then fine-tune on a smaller (though still large) set of robot demonstrations. More recent approaches \cite{shi2025zeromimic,liu2025egozero,lepert2025phantom} explore \textit{zero-shot} learning from human videos without robot demonstrations. Our method differs from \cite{shi2025zeromimic} by avoiding reliance on a goal image — essentially a partial demonstration — and by addressing the human-to-robot embodiment gap. Unlike \cite{liu2025egozero,lepert2025phantom}, we train on in-the-wild human data, and through doing so leverage its advantages. \cite{lepert2025masquerade}, while similar in approach, is not zero-shot since it still requires additional robot demonstrations.

% larger in scale and diversity than the datasets employed in 

% presenting a more challenging problem than curated, author-collected datasets. 

% {\color{blue}make sure to compare with V-JEPA2, which uses more data, different modeling, less actions}

\section{METHODS}
\label{sec_methods}

% {\color{blue} lead with section overview. Zero shot or data primary focus? Lead with main. Integrate with figure 1, etc. Figure captions. Pavan / hardware side?}

%Our policy network is zero-shot as it is \textit{trained} purely over data originating from a human actor performing actions within a kitchen environment.

% During \textit{testing}, the robot is deployed over data featuring a robot manipulator to perform actions within a lab environment, most of the objects lying outside the training distribution of objects. 
% EmbodiSwap: a multi-step data processing pipeline
% This process is dependent on adaptation of human data to robot data. This is a multi-step process involving hand-focused components, as covered in Section {\ref{subsec_data_preparation}}. 
Here we detail our method for zero-shot robot imitation learning from human video. Our model implementation is centered on a fine-tuning of a pre-trained V-JEPA model, the process of which we detail in Section {\ref{subsec_model_architecture}}. This is dependent on EmbodiSwap: a multi-step data processing, as covered in Section {\ref{subsec_data_preparation}}. This involves overlaying robots on human data ({\ref{subsubsec_robot_compositing_process}}), which in turn relies on mapping between human and robot effectors ({\ref{subsubsec_robot_pose_retargetting}}), producing targets for training ({\ref{subsubsec_ground_truth}}), and re-segmenting human data for robot imitation learning ({\ref{subsubsec_manipulation_boundaries}}). Finally, the training of this setup over this data is covered in Section {\ref{subsec_training_deployment}}.

% where  are utilized to blend the contents of This rendering is superimposed over the inpainted scene produced by OmniEraser - 

% We subsequently pass each frame 

\subsection{Model Architecture}
\label{subsec_model_architecture}
% {\color{blue} (More closely integrate with Figures 1}

Here we describe our method's data and model flow as illustrated in Figure \ref{figure_architecture}.

We primarily leverage 2 components of the V-JEPA backbone architecture from \cite{bardes2023v}. V-JEPA originally consists of three components: i) an x-encoder (\textbf{Encoder} within Figure \ref{figure_architecture}) subnetwork that ingests a masked input video and produces an embedded vector for each non-masked video token, ii) a predictor (\textbf{Predictor} within Figure \ref{figure_architecture}) subnetwork that ingests the embedded vectors from the x-encoder along with a set of positional mask tokens each of which corresponds to the masked portion of video, and predicts feature targets for the input video, and iii) a y-encoder subnetwork that takes the entire video and produce deep feature targets. For how V-JEPA is pre-trained, we refer readers to Section 3 of the V-JEPA paper \cite{bardes2023v}.

% Essentially, the encoder-predictor belonging to Figure \ref{figure_architecture} belong to the x-encoder and predictor open-sourced from [], and we discard the y-encoder subnetwork for most experiments (but keep it for the ablation study within Section X).

% % As shown in Figure \ref{figure_architecture}, the encoder-predictor we adopt belong to the x-encoder and predictor open-sourced from [], and we discard the y-encoder subnetwork.

The input to the V-JEPA policy network consists of a robot composited image $I^*_0$ (whose production is described in Section \ref{subsec_data_preparation}), $M_{1:T}$ (the positional mask tokens corresponding to frames $I_{1:T}$), and optional proprioception and action location inputs $p_0$ and $l_0$, respectively. Image $I^*_0$ is fed into the V-JEPA encoder, producing an embedding of the image $I^*_0$. The embedding is fed to the predictor, which uses $M_{1:T}$ to predict an encoded representation of the video $I^*_{0:T}$. Separately, the proprioception input $p_0$ and action location input $l_0$ are encoded by fully connected layers. The output of the predictor as well as the encoded representations of $p_0$ and $l_0$ are concatenated and then fed into attention probe $C$ consisting of two cross attention layers followed by 2 self attention layers. The final output is fed into a fully connected layer which produces a single relative pose vector, where the training signal is derived as per Section \ref{subsec_data_preparation}. 

%Our network produces a single end-effector transform prediction, as opposed to the more conventional \textit{sequence} of transforms that would be predicted by architectures such as {\color{blue} [][]}.

% (3D relative translation, and 6D relative rotation taken from the first two columns of the homogenous rotation matrix, relying on unit normalization to derive the third column)

% The video input (16x256x256) is processed by V-JEPA's encoder and predictor networks, and the proprioction input (4x4 transform) and action location input (1x3 point) are processed with separate fully connected layers. The video, proprioception, and action location tokens are concatenated and fed into 

% {\color{blue} Describe Robot inpainting (see Section X)}
% {\color{blue} 3D hand comparison, Describe training signal}

% \subsection{Data Preparation}
\subsection{EmbodiSwap}
\label{subsec_data_preparation}
% {\color{blue} Overview this subsection}
% Human videos are abudant and easy to collect at scale

% Egocentric video contains a diverse set of cues - particularly relevant to robotic manipulation is the motion of the hands and the motion of the objects of interaction. While both in theory are desirable, extracting the modeling the 3D motion of the objects of interaction poses a number of difficulties {\color{blue} []} and is considerably more error-prone than estimating the 3D motion of the hands. As such, we rely primarily on hand motion.

% The motions of hands and motions of objects are highly correlated. Objects and hands move in synergy. We provide hand derived locations as the hand can be tracked more reliably (object occlusion, appearance variety, etc.). 

% Centering the importance of hands we substitute human hands for robot manipulators for video clips (Section \ref{subsubsec_robot_compositing_process}). This data is then used as input in training the V-JEPA model, with a derived training signal described in Section \ref{subsubsec_ground_truth}. As temporal action boundary annotations in action datasets are ambiguous and inconsistent [] and ill-suited for robot manipulation learning, we describe the re-annotation of the existing annotations in Section \ref{subsubsec_manipulation_boundaries}.

% Particularly relevant to robotic manipulation are the motions of the hands and the objects of interaction.

Human videos are abundant and easy to collect at scale. Furthermore, egocentric video in particular contains a diverse set of cues. The motions of the hands and the objects of interaction are of particular relevance to robotic manipulation. While both are desirable in principle, extracting and modeling the 3D motion of objects poses a number of difficulties and is considerably more error-prone than estimating the 3D motion of hands. Accordingly, we rely primarily on hand motion.

% The motions of hands and objects are highly correlated; they move in synergy. We therefore use hand-derived locations, as hands can be tracked more reliably (e.g., object occlusion, appearance variability). 

Centering the importance of hands, we substitute human hands with robot manipulators in video clips (Section \ref{subsubsec_robot_compositing_process}). The resulting data are then used to train the V-JEPA model with a derived training signal (Section \ref{subsubsec_ground_truth}). Because temporal action boundary annotations in existing action datasets are often ambiguous and inconsistent \cite{alwassel2018diagnosing} and ill-suited for robot manipulation learning, we re-annotate these boundaries (Section \ref{subsubsec_manipulation_boundaries}).

% annotator action boundaries include extraneous sub-actions and exclude essential motions within the action,

% Low inter-annotator consistency and very often exclude frames 

% Given primacy to action categories, rather than action boundaries.

\begin{figure}[t!]
    \includegraphics[width=0.485\textwidth]{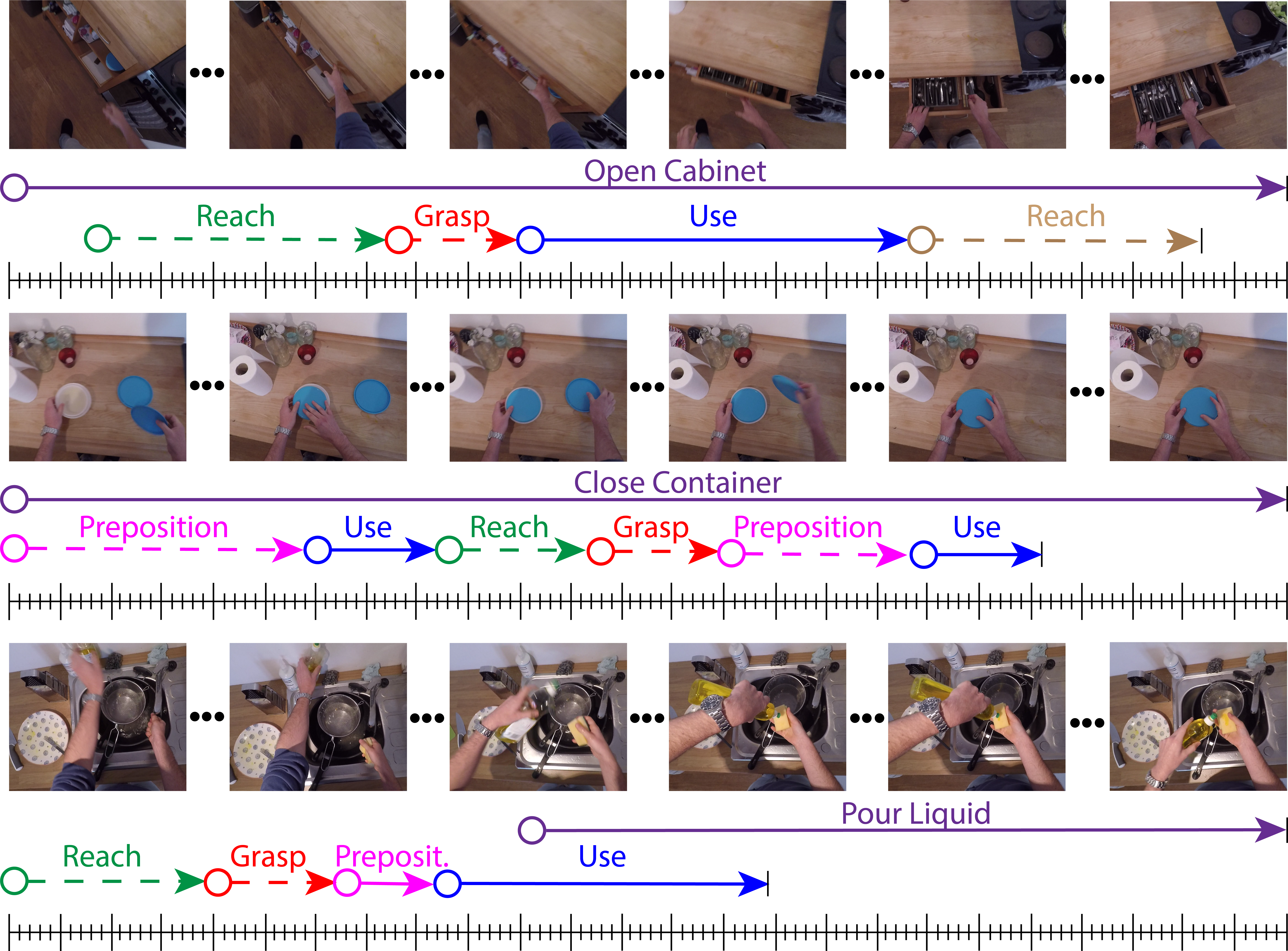}
    \caption{
\textbf{Action and sub-action boundaries:} A visualization of 3 example sequences taken from EPIC Kitchens. The top row within each example corresponds to cropped RGB frames. The {\color{violet} purple} arrow in the middle row corresponds to the ground truth annotated action and its temporal boundaries as provided by EPIC Kitchens. The ({\color{red} red}, {\color{green} green}, {\color{blue} blue}, {\color{magenta} pink}, and {\color{tan} tan}) arrows in the third row correspond to the sequences of Therblig sub-actions, and their temporal boundaries. Both solid and dashed arrows indicate the temporal extend of Therblig sub-actions. A dashed arrow indicates that the Therblig sub-action is extraneous, and a solid arrow indicates that the associated sub-action clip is used in training our system. We release our annotations publicly \href{https://drive.google.com/drive/folders/1-UUywelBCOe-E_ErpoaAHQa4dgjq6AfH?usp=sharing}{here}.}

    \label{figure_data_annotation_illustration}
\end{figure}

%odeling action around Therblig sub-action provides tremendous benefit, as demonstrated empirically in Section \ref{sec_experiments} 
\subsubsection{Robot Compositing Process}
\label{subsubsec_robot_compositing_process}

% {\color{blue} Center of Figure \ref{figure_data_generation_flow}}

% We here detail our process of superimposing robot manipulators over humans in video clips, as detailed in Figure \ref{figure_data_generation_flow}.

Given a clip featuring a human actor performing object manipulation, we wish to transform the RGB frames of that clip in such a fashion that they depict a scene featuring a robotic actor mimicking the motions of the human actor. We detail our process - applied framewise - for this substitution in Figure \ref{figure_data_generation_flow}. The input RGB frame is fed into: 1) Body Segmentation, 2) 3D hand Extractor, and 3) Depth Model. The human is removed by 4) Image Inpainting, and the robot is overlaid by 5) Render and Blend

%performing object manipulation. We detail our process for this substitution in Figure \ref{figure_data_generation_flow}

We extract the 3D hand trajectory of the hand of interaction, using HaWoR \cite{zhang2025hawor}, a 3D reconstruction method that jointly reconstructs the 3D hands and camera pose for each frame. We also feed the entirety of the clip to the frame-wise depth model UniDepthV2 \cite{piccinelli2025unidepthv2}, a network that takes frames along with intrinsics and produces metric valued depth estimates.

Additionally, each frame is fed to a Body Segmentation network that produces a binary mask capturing all pixels that belong to the body of the actor. To generate a high resolution body segmentation mask, we utilize SAM2 \cite{ravi2024sam}, prompted with points belonging to the body from a coarse segmentation of the original image in low-resolution (provided by a network trained over VISOR \cite{darkhalil2022epic}). The binary segmentation output in addition to the original frame are then passed into OmniEraser \cite{wei2025omnieraser}, a high-resolution Image Inpainting network that produces an image of the scene absent of the actor and their effects. 

% Finally, we pass each frame of the original clip to a depth estimation network, producing depth maps (in meters).

The output of these components are then blended, and a final image render in Render and Blend: We re-target the 3D hand extracted from each frame into a gripper pose vector (detailed in Section \ref{subsubsec_robot_pose_retargetting}). We then render an RGB-D synthetic robot at the derived gripper pose using Pybullet's IK. The depth maps belonging to the scene are pixel-wise compared to the rendered depth map belonging to the synthetic robot, where we blend the corresponding inpainted image and the synthetic robot image by selecting for pixels with the smaller depth values between the two maps, yielding an image of a robot manipulator in place of the original human actor.

We compare the scene’s depth map to the robot’s rendered depth map, pixel by pixel. At each pixel, we choose the source with the smaller (nearer) depth to blend the inpainted scene and the robot render. The final result is an image where the robot manipulator replaces the original human actor, while allowing for occlusions on top of the robot manipulator (e.g., by the object).

\subsubsection{Gripper Pose Re-Targetting}
\label{subsubsec_robot_pose_retargetting}
\begin{figure}[t!]
    \includegraphics[width=0.40\textwidth]{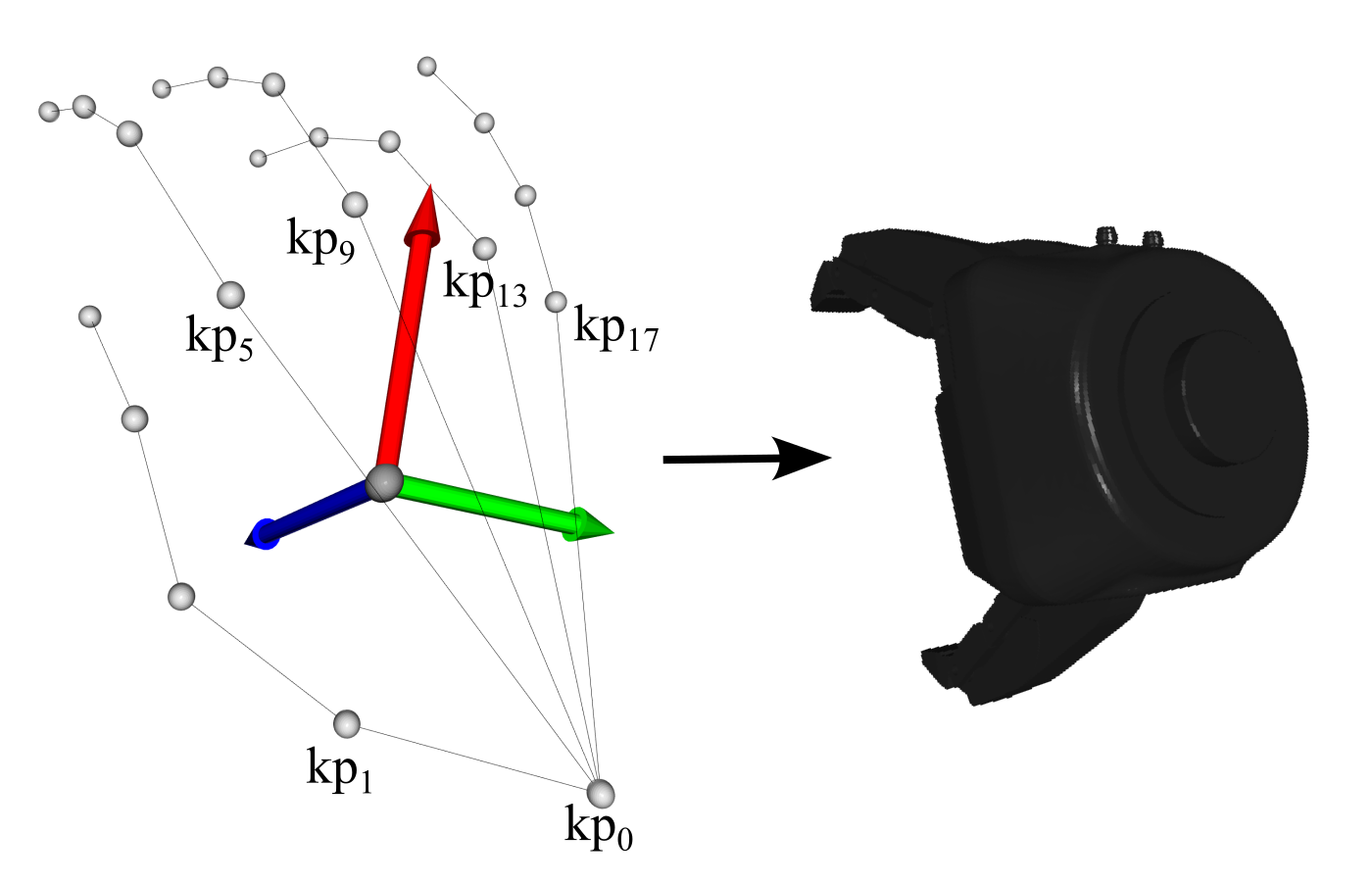}
    \centering
    \caption{
    \textbf{Human Hand Pose to Gripper Pose Re-targeting:} 
    On the left we extract the MANO joint positions of the human hand. From this we extract a 6-DOF hand pose. We then align the robot gripper - on the right - to conform to this derived 6-DOF hand pose.}
    
    % Left — extracted MANO hand joint positions; Right — corresponding gripper pose obtained using our re-targeting scheme.}
    % \textbf{Human Hand Pose to Gripper Pose Re-targeting:} Left — extracted MANO hand joint positions; Right — corresponding gripper pose obtained using our re-targeting scheme.}
    \label{figure_retargetting}
\end{figure}

3D human hand skeletons, as shown in Figure \ref{figure_retargetting}, extracted from video frames cannot be directly applied to a robot due to the embodiment gap. To ensure that the robot videos generated from human demonstrations exhibit physically plausible gripper poses, we use MANO \cite{romero2022embodied} parameters predicted by the HaWoR model in modeling the human hand and subsequently re-target the gripper pose.

% \textcolor{tan}{
% Our re-targeting scheme supports both two and three-finger grippers. Given the MANO \cite{romero2022embodied} parameters, we compute the $21$ joint points $kp_{i}$, $i \in [0,20]$. The gripper center is defined as the palm center, 
% $\mathbf{G}_c = \tfrac{1}{5}(kp_{1}+kp_{5}+kp_{9}+kp_{13}+kp_{17})$. The gripper $Z$-axis is aligned with the normal to the palm, 
% $\mathbf{G}_z = (kp_{5}-kp_{0}) \times (kp_{17}-kp_{0})$. The gripper $X$-axis is aligned with the vector from the first thumb joint to the centroid of the first joints of the remaining fingers, 
% $\mathbf{G}_x = \tfrac{1}{4}(kp_{5}+kp_{9}+kp_{13}+kp_{17}) - kp_{1}$. For the $Y$-axis, we set 
% $\mathbf{G}_y = \mathbf{G}_z \times \mathbf{G}_x$ for the right hand and 
% $\mathbf{G}_y = -\mathbf{G}_z \times \mathbf{G}_x$ for the left hand.  
% The resulting gripper pose $\mathbf{T}_g$ is used both for in-painting the robot and as the regression target for our model.}

Our re-targeting scheme supports both two and three-finger grippers. Given the MANO parameters, we compute the $21$ joint points $kp_{i}$, $i \in [0,20]$. The gripper center is defined as the human palm center, 
$\mathbf{G}_c = \tfrac{1}{5}(kp_{1}+kp_{5}+kp_{9}+kp_{13}+kp_{17})$. The gripper orientation is set as follows: The gripper $Z$-axis is aligned with the normal to the human palm, 
$\mathbf{G}_z = (kp_{5}-kp_{0}) \times (kp_{17}-kp_{0})$. The gripper $X$-axis is aligned with the vector from the first thumb joint of the human to the centroid of the first joints of the remaining fingers, 
$\mathbf{G}_x = \tfrac{1}{4}(kp_{5}+kp_{9}+kp_{13}+kp_{17}) - kp_{1}$. For the $Y$-axis, we set 
$\mathbf{G}_y = \mathbf{G}_z \times \mathbf{G}_x$ when mapping to the right hand and 
$\mathbf{G}_y = -\mathbf{G}_z \times \mathbf{G}_x$ for when mapping to the left hand.  
The resulting gripper pose $\mathbf{T}_g = \big[ \hat{\mathbf{G}}_x \; \hat{\mathbf{G}}_y \; \hat{\mathbf{G}}_z \; \mathbf{G}_c \big] $ is used both for in-painting the robot and as the regression target for our model.

% The resulting gripper pose $\mathbf{T}_g$ is used both for in-painting the robot and as the regression target for our model.}

% \textcolor{tan}{
% \begin{equation*}
% \mathbf{T}_g =
% \begin{bmatrix}
% \hat{\mathbf{G}}_x & \hat{\mathbf{G}}_y & \hat{\mathbf{G}}_z & \mathbf{G}_c
% \end{bmatrix}
% \end{equation*}
% }

\subsubsection{Ground Truth}
\label{subsubsec_ground_truth}

% In constructing the targets our model is to predict, we pair frame $r_n$ with hand prediction $h_{m}, m > n$ where $h_m$ is the pose of the hand at frame $m$, a certain number of frames ahead of where the hand pose $h_n$ is in frame $n$. The frame distance between $m$ and $n$ is a value that varies from action to action - for actions characterized by rapid hand movement, $m$ is set to be closer to $n$ than for actions characterized by slow movement (e.g. pour).

When building the end-effector targets we train our policy network over, we take each video frame and pair it with the hand pose from a future frame. The look-ahead (in frames) varies by action: we use a shorter offset for rapid hand motions (e.g. \textit{open}, \textit{close}) and a longer one for slower movements, such as \textit{pouring}. We compute the relative translation and rotation between the hand pose for the current frame and the hand pose from the future frame as a 6D pose.

%9-DOF vector (3D relative translation, and 6D relative rotation taken from the first two columns of the 3x3 rotation matrix, relying on unit normalization to derive the third column). We utilize a 6D rotation representation as opposed to lower-dimensional rotation representations (e.g. 3D Euler angle or 4D quaternion) due to the discontinuity of all rotation representations with less than 5 dimensions as described in [continuity], which is shown empirically to harm learning.

% {\color{blue} (possibly describe the decision to use 9-DOF instead of 6-DOF, the reason being 9-DOF is better for trainability.)}

\subsubsection{Robot Manipulation Action Boundaries}
\label{subsubsec_manipulation_boundaries}

% {\color{blue} (Center on Figure \ref{figure_data_annotation_illustration})}

% As we're learning motion, we require temporal segmentations which align with the motion to be learned. Inconsistent temporal annotations harms the learning of motion learning as different temporal segmentations in effect label different motions.

% A key limitation of adapting existing egocentric action datasets is that their temporal action boundaries are annotated for action classification, often relying on scene cues to the detriment of hand motion \cite{alwassel2018diagnosing}. For robot learning, where motion dynamics are critical, such boundary labels are suboptimal. 

Different temporal segmentations give different motions. Because we're learning motion, it's important to have precise and consistent temporal segmentation. Existing temporal action boundaries for our data are annotated for action classification which is temporally ill-defined \cite{alwassel2018diagnosing}. To address this, we provide re-annotated temporal boundaries that segment actions based exclusively on hand motion, as shown in Figure \ref{figure_data_annotation_illustration}. To this end we annotate according to the Therblig sub-action ontology. Therbligs are a low-level mutually exclusive contact demarcated set of sub-actions. For a depiction of Therbligs and their usage, see Figure 1 within \cite{dessalene2023therbligs}. In addition to the revised action boundaries, annotators are to indicate the dominant hand of interaction which is missing from existing egocentric datasets. 

\subsection{Training and Deployment}
\label{subsec_training_deployment}

% {\color{blue} (Make (passing?) reference to figure 2)}

% When training the network, rather than train over a sequence of robot-composited frames, we instead duplicate the last image of the sequence temporally, feeding the concatenated stack to the network. 

% The lack of temporal character to the fed input performs marginally better as we show in the ablations within Section X.

After running the robot compositing process over the human images, we arrive at the robot images to be fed into the V-JEPA network. These are paired with the corresponding future hand poses derived in Section \ref{subsubsec_ground_truth}. The V-JEPA network when fine-tuned over action classification tasks typically ingests video input \cite{bardes2023v} - we observe that the training of V-JEPA over video input marginally hurts performance as compared to a single frame input, as shown in the comparisons within Table \ref{tab:tableX_backbones}. For more training details, see Section \ref{sub_training_details}.

% During deployment of the fine-tuned network in the real-world, something that arises is the difference in appearance between the robot embodiment rendered in simulation as captured in the data, and the robot embodiment in the real world.

We deploy our trained policy network in the real-world as shown in Figure \ref{figure_deployment}. There is a difference in appearance between the robot embodiment as rendered in our simulated training data, and the physical robot platform we evaluate over during deployment in the real world. To circumvent this we use the joint angles of the physical robot and render a synthetic RGB image of the physical robot. We overlay this synthetic image over the input image, in a process analogous to that shown in Figure \ref{figure_data_generation_flow}. However, we observe that empirically this substitution is largely unnecessary - the model predictions change only marginally due to this substitution. A square crop from the synthesized image is fed to V-JEPA. The policy network then produces as output a relative transform for the robot's end effector to follow.

{\makeatletter
 \let\orig@makecaption\@makecaption
 \long\def\@makecaption#1#2{%
   \footnotesize\normalfont
   \setlength{\parskip}{0pt}%
   \renewcommand{\baselinestretch}{1}\selectfont
   \vskip\abovecaptionskip           % <-- space above caption
   \noindent\textbf{#1.}\enspace #2\par
   \vskip\belowcaptionskip           % <-- space below caption
 }
\begin{table*}[!t]
\centering

% --- table body (scoped styles so they don't leak into caption) ---
\begingroup
\setlength{\tabcolsep}{4pt}\renewcommand{\arraystretch}{1.05}\scriptsize
\resizebox{\textwidth}{!}{%
\begin{tabular}{l c c c | cc | cc | cc | cc | c | cc}
\toprule
\multirow{2}{*}{Model} & \multirow{2}{*}{Supervision} & \multirow{2}{*}{\makecell[c]{Pretrained\\Modality}} & \multirow{2}{*}{Data type} &
\multicolumn{2}{c|}{Open} & \multicolumn{2}{c|}{Close} & \multicolumn{2}{c|}{Pour} &
\multicolumn{2}{c|}{Cut} & \multicolumn{1}{c|}{Place} & \multicolumn{2}{c}{Composite} \\
 &  &  &  & Trans & Rot & Trans & Rot & Trans & Rot & Trans & Rot &   & Trans & Rot \\
\midrule
ResNet-50            & supervised  & image & non-robot & 0.119 & 0.394 & 0.115 & 0.412 & 0.083 & 0.358 & 0.061 & 0.250 & 0.167 & 0.109 & 0.354  \\
MAE ViT-B/16  \cite{he2022masked}       & self-supervised  & image & non-robot & 0.089 & 0.342 & 0.095 & 0.342 & 0.071 & 0.312 & 0.057 & 0.247 & 0.147 & 0.092 & 0.311  \\
DINOv2 \cite{oquab2023dinov2}               & self-supervised & image & non-robot & 0.084 & 0.359 & 0.094 & 0.369 & 0.071 & 0.301 & 0.051 & 0.251 & 0.119 & 0.084 & 0.320  \\
R3M \cite{nair2022r3m}                  & supervised & image & non-robot & 0.109 & 0.401 & 0.116 & 0.37 & 0.099 & 0.415 & 0.061 & 0.262 & 0.178 & 0.113 & 0.362 \\
ViP \cite{ma2022vip}              & self-supervised & image & non-robot & 0.110 & 0.376 & 0.119 & 0.444 & 0.106 & 0.399 & 0.065 & 0.345 & 0.192 & 0.118 & 0.391  \\
VC-1 \cite{majumdar2023we}                & self-supervised & image & non-robot & 0.101 & 0.372 & 0.104 & 0.385 & 0.080 & 0.326 & 0.0584 & 0.249 & 0.139 & 0.096 & 0.333  \\
Octo \cite{team2024octo}           & supervised  & image & robot & 0.147 & 0.415 & 0.138 & 0.392 & 0.085 & 0.363 & 0.062 & 0.273 & 0.220 & 0.130 & 0.361  \\
RoboFlamingo \cite{li2023vision}          & supervised  & image & robot & 0.139 & 0.430 & 0.140 & 0.391 & 0.083 & 0.371 & 0.599 & 0.340 & 0.169 & 0.226 & 0.383  \\
$\bm{\pi}_0$ \cite{black2024pi0}                  & self-supervised & image & robot & 0.102 & 0.361 & 0.104 & 0.36 & 0.089 & 0.336 & 0.054 & 0.235 & 0.160 & 0.102 & 0.323  \\
ResNet-3D                 & supervised  & video & non-robot & 0.110 & 0.416 & 0.109 & 0.373 & 0.095 & 0.315 & 0.048 & 0.331 & 0.195 & 0.111 & 0.359 \\
Hiera-L \cite{ryali2023hiera}               & self-supervised  & video & non-robot & 0.092 & 0.363 & 0.104 & 0.399 & 0.071 & 0.296 & 0.050 & 0.290 & 0.155 & 0.094 & 0.337  \\
V-JEPA + T (ViT-L)              & self-supervised         & video  & non-robot     & 0.085 & 0.338 & 0.081 & 0.280 & 0.065 & \textbf{0.239} & 0.049 & 0.239 & 0.101 & 0.076 & \textbf{0.274}  \\
V-JEPA (ViT-H)            & self-supervised         & video & non-robot      & 0.086 & \textbf{0.319} & \textbf{0.080} & 0.334 & \textbf{0.061} & 0.256 & \textbf{0.042} & \textbf{0.236} & \textbf{0.094}  & \textbf{0.073} & 0.286 \\
V-JEPA (ViT-L)                & self-supervised & video & non-robot & \textbf{0.082} & 0.335 & 0.082 & \textbf{0.269} & 0.069 & 0.260 & 0.045 & 0.236 & 0.099 & 0.076 & 0.275  \\
\bottomrule
\end{tabular}}
\endgroup

% --- caption (put after body if you prefer bottom; move above for IEEE default top) ---
\caption{\textbf{Pre-training Method Comparison:} Here we evaluate the utility of different pre-training methods trained over EmbodiSwap data for the task of forecasting end-effector pose, over the following actions: \textit{Open}, \textit{Close}, \textit{Pour}, \textit{Cut} and \textit{Place}. We define pre-training method according to Supervision, Modality, and Data Type categories. Supervision is broken into classification (whether it be object or action classification), self-supervision (where the training signal is derived from input), and action prediction (regression of end-effector pose). Modality can be either image or video. Data type is either robotic data, non-robotic data, or a mix of both. We present the final model performance across 5 actions independently, as well as in aggregate. For each, performance is divided into translation prediction error (in meters), and rotation prediction error (unitless). For models pretrained over images, we use images alone. For models pretrained over videos (except VJEPA), we use images cloned across the temporal dimension. For V-JEPA + T we use video sequences. For V-JEPA (ViT-H and ViT-L) we feed input as per Figure \ref{figure_architecture}.}

% Within the models, V-JEPA uses a ViT-L backbone unless mentioned otherwise, and V-JEPA + T corresponds to using videos as opposed to single images + positional masks.

% For models pretrained over images, we use images alone. For models pretrained over videos (except VJEPA), we use images cloned across the temporal dimension. For V-JEPA + T we use video sequences. For V-JEPA (ViT-H and ViT-L) we feed input as per Figure \ref{figure_architecture}.

% For V-JEPA (ViT-H and ViT-L), we use a single frame in addition to a stack of positional mask tokens as in Figure \ref{figure_architecture}. 

\label{tab:tableX_backbones}
\end{table*}

 \let\@makecaption\orig@makecaption
 \makeatother}

\section{EXPERIMENTS}
\label{sec_experiments}

% Here we detail our experimental evaluation. Data is characterized in Section \ref{} which goes into the actions we model

Here we detail our experimental evaluation. Data is characterized in Section \ref{sub_data}, and training details are provided in Section \ref{sub_training_details}. We provide an assessment over pre-training methods in Section \ref{sub_pre_training_method_evalutions}, demonstrating the superiority of feature-level video prediction pretraining over other pre-training methods. In Section \ref{sub_real_world_experiments} we present evaluations performed using a physical platform in our lab.

\subsection{Data}
\label{sub_data}
% There are a plethora of actions that can be modeled over egocentric video - however, egocentric actions are typically long-tailed []. 

We model actions most frequently labeled within the action datasets - in this case, \textit{placing}, \textit{opening}, \textit{closing}, \textit{pouring} and \textit{cutting}. We source videos from $3$ popular egocentric datasets - EPIC Kitchens \cite{damen2022rescaling}, HOI4D \cite{liu2022hoi4d} and Ego4D \cite{grauman2022ego4d}. We source all five actions from EPIC Kitchens 2020, \textit{placing}, \textit{opening} and \textit{closing} actions from HOI4D and \textit{pouring} and \textit{cutting} videos from Ego4D.

% {\color{blue} (if there's space: characterize the datasets, their applicability to this project, and specifically the differences between the datasets)}

% We amend the temporal action boundaries as per Figure \ref{figure_data_annotation_illustration} as provided action boundaries are associated with action classification, and thus are associated more with scene and object cues than with hand motion. When amending the action boundaries within the existing action labels, we prompt annotators to re-annotate the boundaries of the action clips according to the Therblig ontology.

When prompting annotators to re-annotate boundaries of action clips as in Section \ref{subsubsec_manipulation_boundaries}, we define actions to begin when the tool object is \textit{grasped}, and terminate before the end of the \textit{use} operation, which corresponds to the post-condition of the action being satisfied (e.g. door finishes \textit{opening}/\textit{closing}, cuttable object is sliced, etc).

%Frames belonging to the original action boundaries that contain extraneous Therbligs (e.g. \textit{grasp}, \textit{release}, \textit{wait}, etc) are not included within the amended action boundaries. 

%We also instruct annotators to exclude frames containing dexterous in-hand manipulation motions.

% Open and close both begins after grasp, close begins after grasp, pour begins at preposition and ends after use

We do training and evaluation over the different actions independently, as opposed to pooling the different actions into a single dataset. For the \textit{open}, \textit{close}, and \textit{cut} action, we make no use of the action location as input to the network, feeding a placeholder instead. For the \textit{pour} action, we feed a noisy estimate of the action location as input so as to encourage the network to be rely on visual cues for the \textit{pouring} action. For all actions, except the \textit{place} action, we include proprioception to the network - for the \textit{place} action, we observe that the network performs better when denied access to proprioception information.

\subsection{Training Details}
\label{sub_training_details}

%After preparing amended action boundaries for each sequence as described in X and running the revised sequences through the data editing process as described in Y, we construct datapoints 

We resize images from the original resolutions to a fixed resolution of $1080 \times 1920$. During training we apply random crops (scales range from $0.4$ to $0.99$) and random horizontal flips, which we find helps generalization for all actions (non-spatial augmentations such as color augmentations tend to hurt generalization). Cropped images are then resized to resolution $224 \times 224$ and fed into the policy network. 

For the results reported in Section \ref{sub_pre_training_method_evalutions}, we freeze all layers except the attention probe. For the results reported in Section \ref{sub_real_world_experiments} (and in accordance with Figure \ref{figure_architecture}, we only freeze the encoder, training the predictor as well as the attention probe.

We train our policy network for $40$ epochs with a batch size of $32$, an initial learning rate of $1e-3$, and a cosine scheduler that ends the learning rate at $1e-7$. We also adopt a weight decay of $0.01$. Because we keep the V-JEPA encoder backbone frozen, we are able to achieve the memory requirements of the single RTXA5000 GPU we use for training.

% We observe that geometric augmentations improve generalization, 

% whereas color augmentations hurt generalization. 

% \begin{table}[htbp] % [h]ere, [t]op, [b]ottom, [p]age
%     \centering
%     \caption{Example Table of Data}
%     \label{tab:example_data}
%     \begin{tabular}{lccr} % l: left, c: center, r: right alignment
%         \toprule % Top rule from booktabs package
%         \textbf{Header 1} & \textbf{Header 2} & \textbf{Header 3} & \textbf{Header 4} \\
%         \midrule % Middle rule from booktabs package
%         Row 1, Col 1 & Row 1, Col 2 & Row 1, Col 3 & Row 1, Col 4 \\
%         Row 2, Col 1 & Row 2, Col 2 & Row 2, Col 3 & Row 2, Col 4 \\
%         \bottomrule % Bottom rule from booktabs package
%     \end{tabular}
%     \caption{Caption!!!!!!!!as asdf asdf aasdfasdfasdfa asdf asdf asd asd asd asd asdf asd asd fasd asd asdf asdf asf asdf sdf asdf}
% \end{table}

% \subsection{Visual Backbone Evaluations}
\subsection{Pre-training Method Comparisons}
\label{sub_pre_training_method_evalutions}

% {\color{blue} (Structure around Table X)}

% A column for model name

% Plus: table column:
% supervision (classification, actions, self-supervised)
% modality (image or video)
% data type (non-robot, robot, both)

% Eval columns (report both translation/rotation):
% action1
% action2
% action3
% action4
% action5

% Plus composite eval colum

% table row:
% model (20+)

% Put dummy numbers for each reported number 

% reported performances: translation error, rotation error

% {\color{blue} (TODO: consider the particular phrasing for our task: "end-effector pose prediction"?)}

Here we wish to evaluate the impact of the selection of different pre-training methods for the final task of end-effector pose prediction. A pre-training method is categorized based on which supervision category, modality type, and data type it employs, as specified in Table \ref{tab:tableX_backbones}. This is important to explore as it determines the extent to which advancements in vision can be transferred directly into robot learning. The data input are the robot images produced by EmbodiSwap, and the ground truth are their paired future poses.

% without any need for pre-training paradigms specific to robotics.

% Here we wish to evaluate the extent to which the choice of visual backbone determines the performance of the network in forecasting future end-effector movements. 

% that belong into one of the following categories: 1) image networks pre-trained over Internet-scale data, 2) video networks pre-trained over Internet-scale data, 3) image networks pre-trained over robot action prediction, 3) image networks pre-trained over robot action prediction, 4) image networks pre-trained over masked image reconstruction, 5) video networks pre-trained over masked video reconstruction, 6) image networks pre-trained over feature-level image prediction, and 7) video networks pre-trained over feature-level video prediction. We also indicate whether the network being considered was trained on a fully supervised or a self-supervised learning task.

\begin{figure}[b!]
    \includegraphics[width=0.485\textwidth]{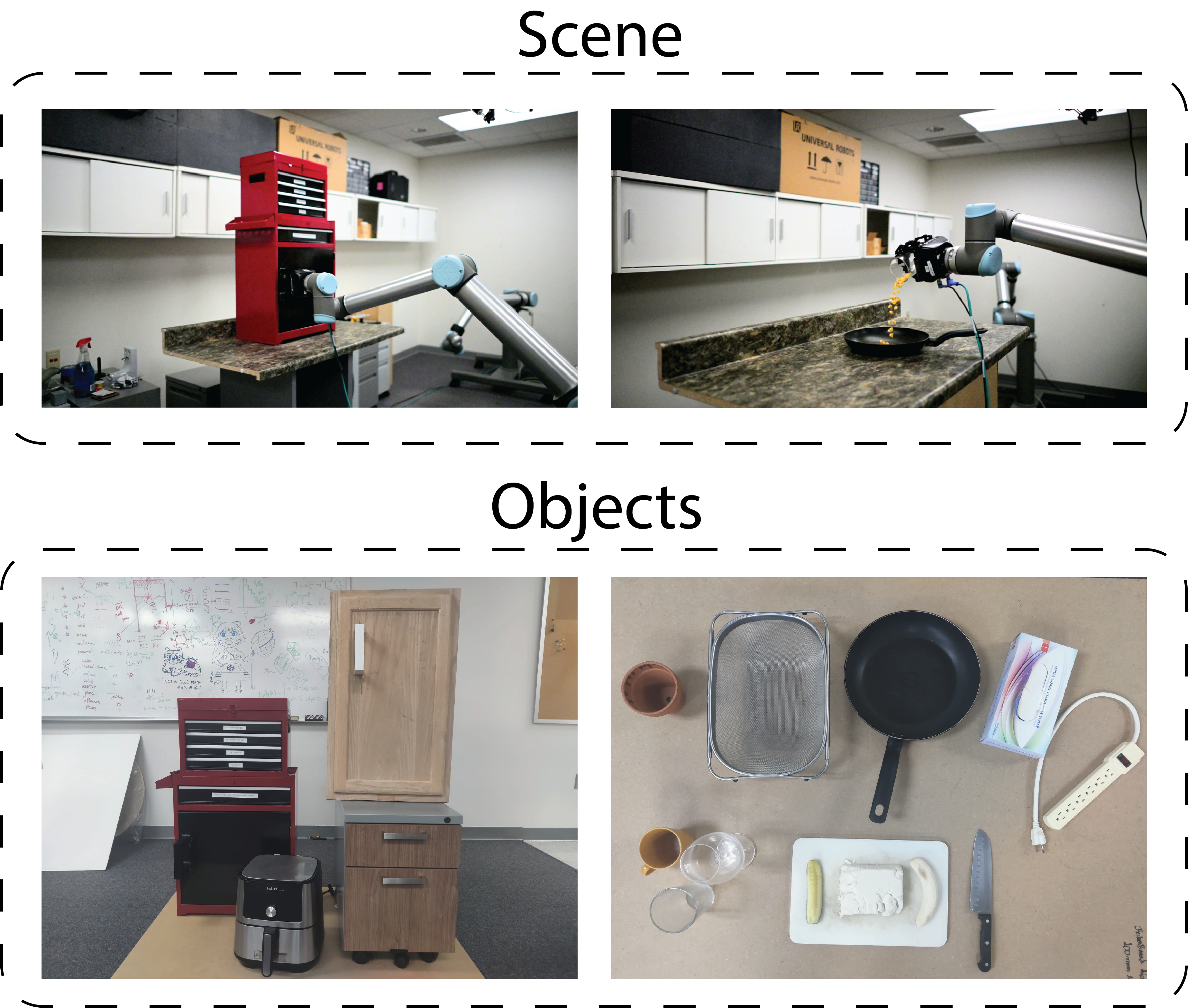}

% \textbf{The robot (UR10) we deploy for each of the following actions that take place in the real-world: open, close, pour, cut and place. The top row depicts on the left side the robot opening the toolbox and on the right side pouring the contents of a cup onto a pan - the bottom row depicts on the left side the objects used for open and close, and on the right side the objects used for cut, place and pour.}}
    \caption{\textbf{Real World Experimental Setup:} We deploy a UR10 robot in the real world to perform each of the following actions: \textit{open}, \textit{close}, \textit{pour}, \textit{cut} and \textit{place}. The top row depicts \textit{opening} a toolbox on the left side, and \textit{pouring} the contents of a cup into a pan on the right side - the bottom row depicts the objects used for \textit{open} and \textit{close} on the left side, and the objects used for \textit{cut}, \textit{place} and \textit{pour} on the right side.}
    \label{figure_realworld_experiments}
\end{figure}

We choose a suite of pre-trained vision backbones for comparison associated with state-of-the-art vision models. In these experiments we follow the convention of \cite{bardes2023v} in swapping out the encoder-predictor subnetworks for the vision backbone of the evaluated model. The attention probe operates over the frozen features produced by the evaluated model, identically as in Section \ref{subsec_model_architecture}. For simplicity, we keep the vision backbones frozen to allow for identical learning settings across experiments, only training the attention probe and cross attention layers. We also do not feed in proprioception or action location inputs so as to focus evaluation on the visual backbone.

% \begin{table*}[!t]
% \centering
% % \caption{Caption}
% \end{table*}

\subsection{Real-World Experiments}
\label{sub_real_world_experiments}

We conduct our experiments on a real-world UR10 robot with a Robotiq gripper controlled using ur\_rtde interface \cite{10871000}. The environment we deploy over is a lab setting with different background and lighting conditions than the kitchen environments the policy network is trained over (illustration in Figure \ref{figure_realworld_experiments}). Of the 5 actions we demonstrate, 2 of the actions (\textit{open} and \textit{close}) are contact rich in that the robot's commanded motion is constrained by the articulated joint of the objects of interaction. For these actions, we adopt a force controller that allows compliance in all 6 degrees of freedom, with the tradeoff of precise control. For the remaining three actions (\textit{pour}, \textit{cut} and \textit{place}) we adopt a position controller, which terminates execution upon collision but gives precision in reaching commanded movements.

% The following are how we evaluate an action's outcome: 1) open succeeds when revolute door exceeds 65 degrees or translational door exceeds X percent displacement, 2) close only succeeds when contact is made between the cover and the base, 3) for pour we count the number of pieces that fall into the vessel, 4) cut only succeeds when the knife slices a target object, and 5) place only succeeds when the body of the object intersects with the designated action location. We make the place action more challenging by only placing large objects where the robot hand is displaced far away from the centroid of the object. This forces the network to model the spatial relations between the object being placed and the target location (as opposed to modeling the easier spatial relations between the hand and the target location as would be done when placing small objects). We randomize positions and orientations of all objects over each of the trials. For a visual depiction of all objects we evaluate on, see Figure \ref{figure_realworld_experiments}. We report our results in Table \ref{real_world_eval}.

We evaluate action outcomes using the following criteria: (i) \textit{open} succeeds when a rotational door rotates beyond 65° or if a translational drawer extends more than $80\%$ of its maximum range; (ii) \textit{close} succeeds only when the cover makes contact with the base; (iii) \textit{pour} succeeds when over $80\%$ of the styrofoam pieces (we avoid liquids) fall into the receiving vessel; (iv) \textit{cut} succeeds only when the knife slices through the target object; (v) \textit{place} succeeds when the object makes contact with the point belonging to the designated target location.

\begin{table}[!t]
\centering
% \caption{\textbf{Real World Evaluation:} Comparison of $3$ different models - First row being $\pi_0$-FAST finetuned over $30$ in-lab collected demonstrations (few-shot), second row being the $\pi_0$-FAST network trained over the data produced by EmbodiSwap alone (therefore zero-shot), and the third row being our proposed framework discussed in Section \ref{sec_methods}. We note that the "*" entry for $\pi_0$-FAST indicate that the model performed extremely poorly on the held-out validation set of in-lab collected demonstration, and so $\pi_0$ was not physically deployed on these actions.}

\caption{\textbf{Real World Robot Evaluation:} Comparison of $3$ different models (Section \ref{sub_real_world_experiments} for details). 30 refers to training over 30 in-lab demonstrations. ES refers to EmbodiSwap. \textbf{Our} method is V-JEPA over ViT-L with EmbodiSwap data.}
% We note that the "-" entries for $\pi_0$ indicate that the model is very unlikely to achieve any successes in the real-world due to extremely poor performance on the held-out validation set, and so $\pi_0$ was not physically deployed on these actions.

    \label{real_world_eval}
\setlength{\tabcolsep}{3pt}
\renewcommand{\arraystretch}{1.15}
\begin{tabular}{lcccccc}
\toprule
\textbf{Method} & \textbf{Open} & \textbf{Close} & \textbf{Pour} & \textbf{Place} & \textbf{Cut} & \textbf{All} \\
\midrule
\textbf{$\bm{\pi}_0$ (30)}    & 6/20 & 4/20 & 2/15 & 2/15 & 1/15 & 15/85 \\
\textbf{$\bm{\pi}_0$ (ES)}             & 18/20 & 3/20 & 1/15 & 0/15 & 0/15 & 24/85 \\
\textbf{Ours} & \textbf{19/20} & \textbf{17/20} & \textbf{10/15} & \textbf{10/15} & \textbf{14/15} & \textbf{70/85} \\
\bottomrule
\end{tabular}
\end{table}

To capture the complexity of place, we employ large objects (pan, power strip, box, etc.) for manipulation, and initiate the robot grasp of the object on the object periphery, away from the object centroid. This requires V-JEPA to learn to model spatial relations involving object shape, gripper alignment, and target location. This differs from placing small objects, as small objects can be placed simply by positioning the gripper directly over the target location, requiring neither object nor gripper spatial modeling. 

Across trials of \textit{pour} and \textit{cut} tasks, we randomize the positions and orientations of objects within a rough $120$ cm $\times 60 $ cm planar workspace. For both the \textit{open} and \textit{close} tasks, we randomize the 6D grasp pose of the gripper to evaluate each method’s robustness to variations in grasp configuration. Figure \ref{figure_realworld_experiments} provides a visual overview of the evaluated objects, and quantitative results are reported in Table \ref{real_world_eval}.

Looking for comparisons, the most directly comparable method would be \cite{shi2025zeromimic} but they do not release checkpoints or inference code. Instead, we compare two variants of the $\pi_0$ model \cite{black2024pi0} with \textbf{Our} proposed approach employing a V-JEPA ViT-L backbone trained over EmbodiSwap data in Table \ref{real_world_eval}. The \textbf{$\bm{\pi}_0$ (30)} baseline refers to the $\pi_0$ network fine-tuned over $30$ in-lab collected demonstrations (we use default training settings, aside from an action horizon of $3$ and $5000$ training steps. The \textbf{$\bm{\pi}_0$ (ES)} baseline refers to the pretrained vision backbone of $\pi_0$ network fine-tuned over the data produced by EmbodiSwap, without any access to in-lab robot demonstrations.

\section{Discussion and Future Work}

% {\color{blue} (Introduce 2-ish paragraphs discussing Table I results). (Some amount of discussion of results) for pre-training comparison}
We compare $13$ different vision backbones on the task of forecasting end effector trajectories associated with egocentric video and report results in Table \ref{tab:tableX_backbones}. V-JEPA (ViT-L) outperforms all vision backbones outside the class of V-JEPA models. The networks trained over large-scale robot data (RoboFlamingo, $\pi_0$, Octo) are among the least competitive, indicating that learning over the distribution of robot trajectories is not helpful in predicting trajectories associated with egocentric video. Curiously, we notice the second most competitive network aside from V-JEPA is DINOV2 \cite{oquab2023dinov2} - which shares a similarity with V-JEPA in that they are both pre-trained over the task of feature-level prediction.% The VC-1 \cite{majumdar2023we}

% \textbf{Ours (V-JEPA)} 
% $ \bm{\pi} $

% and upon qualitative observation observe that the trajectories of Our method appear better formed.

% and upon qualitative inspection this holds even for the instances where the baselines succeeded in performing the action.

% \textbf{$\mathbf{\bm{\pi}_0}$}

% We train \textbf{$\bm{\pi}_0$-FAST} on $30$ in-lab demonstrations per action and compare against the performance of \textbf{Our} method in Table~\ref{real_world_eval}.

We train \textbf{$\bm{\pi}_0$} on $30$ in-lab demonstrations per action and compare against the performance of \textbf{Our} method (V-JEPA over ViT-L with EmbodiSwap data) in Table~\ref{real_world_eval}. We observe that quantitatively \textbf{Our} approach produces better predictions over all actions. Furthermore, upon qualitative observation we observe that the trajectories of \textbf{Our} method appear better formed. We observe that \textbf{$\bm{\pi}_0$} commonly fails even for objects trained over, likely due to too few demonstrations (the authors of \cite{black2024pi0} do not state a minimum, but $30$ appears insufficient). We observe that the performance of \textbf{$\bm{\pi}_0$} is sensitive to end-effector orientation: it succeeds when the end-effector orientation aligns with the required motion for \emph{open}/\emph{close}, and fails otherwise. Training \textbf{$\bm{\pi}_0$} instead over EmbodiSwap data makes \textbf{$\bm{\pi}_0$ (ES)} competitive on \emph{open}, but it still lags on \emph{close}, \emph{pour}, \emph{cut}, and \emph{place} (Table~\ref{real_world_eval}). The main failure modes of \textbf{Our} method are: (i) joint singularities causing shutdowns; (ii) self-occlusions that remove critical visual cues; and (iii) visually plausible but unsuccessful trajectories (e.g., narrowly missing the receptacle when \textit{pouring}). (i) can be mitigated by predicting in robot joint space to avoid singularities and (ii) can be mitigated by using multiple perspectives as input during inference; we leave these to future work.

Results across Tables~\ref{tab:tableX_backbones} and \ref{real_world_eval} support the intuition that lower hand-forecasting error correlates with real-world success: our method (V-JEPA, ViT-L) achieves $34\%/15\%$ lower translational/rotational error than \textbf{$\bm{\pi}_0$} within Table \ref{tab:tableX_backbones} and achieves a $54\%$ higher success rate when deployed in the real world within Table \ref{real_world_eval}. We find this encouraging, as it implies that advancements in the task of forecasting human trajectories are directly transferrable into robot learning.

Future work involves leveraging 3D object modeling, particularly for tasks where in-hand manipulation is required. Another interesting avenue of exploration involves the composition of our trained policies within higher-level cognitive frameworks \cite{beetz2025robot}. There are many possible future directions for this work, and we encourage others to use the released data, code and models.

----------------------------------------

% We also observe \textbf{Our} approach in the third row of Table \ref{real_world_eval} consistently produces predictions that are of higher quality even for the instances where the baseline succeeded in performing the action. 

% We also observe from qualitative inspection that \textbf{Our} approach in the third row of Table \ref{real_world_eval} consistently produces predictions that are of higher quality even for the instances where the baseline succeeded in performing the action. 

% We quantitatively observe that \textbf{Our} approach in the third row of Table \ref{real_world_eval} produces better predictions over all actions. Furthermore, we qualitatively inspection this holds even for the instances where the baseline succeeded in performing the action. 

% \section{CONCLUSIONS}
\label{sec_conclusion}


\printbibliography

\end{document}